%% file: Icra-2018.tex
\documentclass[letterpaper, 10 pt, conference]{ieeeconf}  % Comment this line 

\IEEEoverridecommandlockouts                              % This command is only
\usepackage{graphics} % for pdf, bitmapped graphics files
\usepackage{graphicx}
\usepackage{epsfig} % for postscript graphics
\usepackage{subfigure}
\usepackage{multirow}
\usepackage{booktabs}

\usepackage[font=small,skip=0pt]{caption}

\usepackage{amsmath} % assumes amsmath package installed
\renewcommand{\baselinestretch}{.93}

\usepackage{textcomp,    % for \textlangle and \textrangle macros
            xspace}
\graphicspath{{./Figs/}}

\newcommand{\nr}[3]{$\langle#1 | \mathbf{#2} | #3\rangle$}

\title{\LARGE \bf
A Deep Learning Based Behavioral Approach to \\ Indoor Autonomous Navigation
}

\author{G.~Sepulveda, J.~C.~Niebles, and A.~Soto%
\thanks{G.~Sepulveda and A.~Soto, Dept.~of Computer Science, Pontificia 
Universidad Catolica de Chile,
        {\tt\small (gsepulveda@uc.cl, asoto@ing.puc.cl)}}%
\thanks{J.~C.~Niebles Dept.~of Computer Science, Stanford University,
        {\tt\small (jniebles@stanford.edu)}}%
\thanks{This work was partially funded by FONDECYT grant 1151018}%
}

\begin{document}

\maketitle
\thispagestyle{empty}
\pagestyle{empty}

%%%%%%%%%%%%%%%%%%%%%%%%%%%%%%%%%%%%%%%%%%%%%%%%%%%%%%%%%%%%%%%%%%%%%%%%%%%%%%%%
\begin{abstract} We present a semantically rich graph representation for indoor robotic navigation. 
Our graph representation encodes: semantic locations such as offices or corridors as nodes, and 
navigational behaviors such as enter office or cross a corridor as edges. In particular, our 
navigational behaviors operate directly from visual inputs to produce motor controls and are 
implemented with deep learning architectures. This enables the robot to avoid explicit computation 
of its precise location or the geometry of the environment, and enables navigation at a higher level 
of semantic abstraction. We evaluate the effectiveness of our representation by simulating 
navigation tasks in a large number of virtual environments. Our results show that using a simple 
sets of perceptual and navigational behaviors, the proposed approach can successfully guide the way 
of the robot as it completes navigational missions such as going to a specific office. Furthermore, 
our implementation shows to be effective to control the selection and switching of behaviors. 
\end{abstract}

%%%%%%%%%%%%%%%%%%%%%%%%%%%%%%%%%%%%%%%%%%%%%%%%%%%%%%%%%%%%%%%%%%%%%%%%%%%%%%%%
\section{INTRODUCTION}
Currently, the autonomous navigation system of most mobile robots relies on a fine-grained 
geometric world representation, e.g., a metric map \cite{Thrun:2006}. As a main drawback, this 
type of representation exhibits a strong reliance on low level structural information, such as 
specific geometric configurations of the environment \cite{Araneda:Soto:2007} or low level visual 
cues \cite{Davison:SLAM:2008}\cite{vSLAM-Review:12}, limiting its generality and robustness. This 
is particularly relevant in GPS-denied environments, mainly indoor spaces, where tasks such as 
robot localization and planning/exploration, strongly depend on the underlying representation of 
the environment. As a further disadvantage, the use of a low level representation limits the ability 
of the robot to interact with human users, who usually understand the environment, and refer to it, 
at a higher level of abstraction. Recently, the so-called semantic mapping techniques emerge as a 
step 
forward to improve the previous limitations \cite{Kostavelis:Gasteratos:2015}. These approaches, 
however, are usually built on top of low-level geometric representations, inheriting most of their 
problems \cite{Kostas:ICRA:2017}.  

At a more fundamental level, we believe that current representations for robot navigation do not 
take full advantage of the rich semantic structure behind man-made environments. In particular, most 
man-made environments are designed to facilitate human navigation. Consequently, they are mainly 
composed of navigational structures, such as sideways, corridors, or stairs, that in turn are 
intended to connect meaningful neighboring places, such as houses, rooms, or halls. We hypothesize 
that, by providing robots with suitable abilities to understand the world at this semantic level, it 
is possible to equip them with navigational systems that largely exceed the generality and 
robustness of current approaches. 

In this work, we revisit early approaches based on a graph-based cognitive map of the environment 
\cite{Kuipers:1978}. In particular, for the case of an office building, we propose a 
graph representation similar to the one depicted in Figure 1. In this graph, nodes correspond to 
\textit{perceptual behaviors} which are intended to recognize specific instances of high level 
semantic structures, such as particular offices, corridors, or halls. Edges in the graph 
correspond to \textit{navigational behaviors}, such as cross-a-corridor, leave-an-office, or 
enter-a-room, which are intended to lead the robot as it navigates between 
semantic places. 

Departing from current SLAM approaches 
\cite{Kostavelis:Gasteratos:2015}\cite{Kostas:ICRA:2017}, the proposed representation questions the 
need to keep an explicit estimation of metric or low-level geometrical information about the 
environment. To illustrate this idea, consider the case of a robot that needs to cross a corridor. 
To achieve this goal, the robot does not need to know its precise position within the corridor. 
Instead, it can succeed by just strictly following the corridor until reaching its end. In this 
sense, we advocate for a representation based on a perception for action paradigm, where perception 
is an attentive modality that focuses on the basic needs to 
achieve the current robot goals \cite{Gibson:1986}.  

\begin{figure}
\begin{center}
\includegraphics[width=6cm]{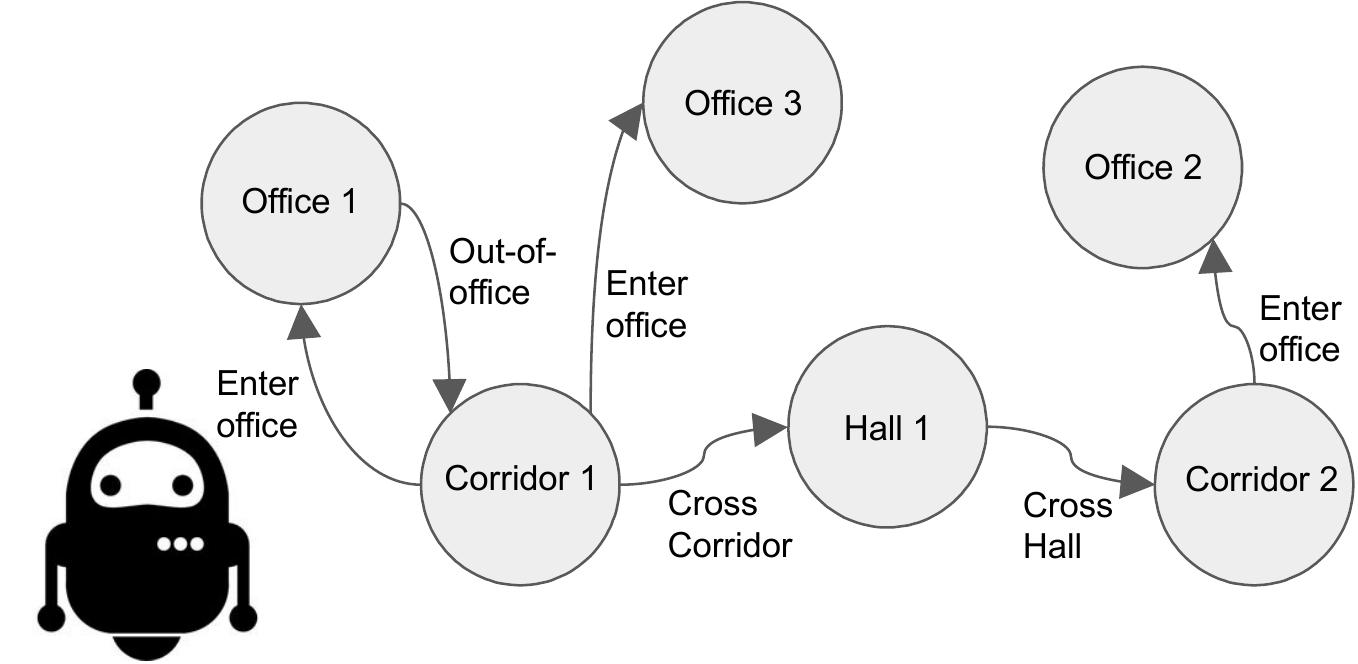} 
\end{center}
\caption{Proposed representation for robot navigation. Nodes 
represent perceptual 
behaviors used by the robot to recognize specific semantic places, such as 
\textit{Office-1} or \textit{Corridor-2}. Edges represent navigational 
behaviors used by the robot to 
move between relevant places, such as 
\textit{leave-an-office} or \textit{cross-a-corridor}.} \label{Fig:Graph}
\end{figure}

The proposed scheme is reminiscent of early behavioral based approaches for 
autonomous robot navigation \cite{Brooks:1986}\cite{Horswill:1993}. These early 
attempts rely on low level sensory information or hand-crafted visual features 
that, at the time, lacked of sufficient robustness to deal with the 
complexities 
of natural environments. Fortunately, current deep learning (DL) 
techniques open a new opportunity to revisit these ideas.  In effect, DL offers 
the possibility to provide a robot with learning skills to acquire suitable 
behaviors by directly experimenting with a real \cite{Bojarski:NVIDIA:2016} or 
a simulated environment \cite{Yuke:EtAl:2017}. 

As a proof of concept, in this work we experiment using virtual environments 
that simulate office building spaces. This facilitates the acquisition of 
training data as well as the evaluation of different aspects of the proposed 
approach. Our main goal is to demonstrate how a robot can navigate in these 
simplified spaces using a set of predefined perceptual and navigational behaviors. Our 
implementation 
follows recent work on visual problem solving using perceptual modules 
\cite{TreborGroup:EtAl:2016} \cite{Justin:EtAl:2017} 
but, in contrast to those works, our planning problem involves a temporal 
dimension that, at execution time, implies the determination of suitable 
switching times between behaviors. Specifically, during an initial exploration 
phase, the robot builds a representation of the environment by recording the 
set 
of behaviors that are being activated. Afterwards, the robot uses this 
representation to solve and execute planning problems, such as going from 
semantic place A to semantic place B. At planning time, this implies selecting 
and sorting a suitable set of behaviors to reach the intended goal. At 
execution 
time, this implies the correct identification of places and transition times between 
the activation of navigational behaviors.

As a distinguishable feature, this planning scheme resembles the semantic 
representation used by a human to guide a stranger in a new environment. 
Indeed, when presenting a new environment, the usual directions provided by 
a human are related to the activation of perceptual and navigational behaviors, such 
as \textit{``cross the hall, take the corridor to the right, and walk until you find a large 
door''}. In this sense, during navigation, the activation of the intended
actions (navigational 
behaviors) is usually associated to the identification of specific 
perceptual cues (perceptual 
behaviors). These are the guiding observations behind our proposed approach. 
In particular, we make the following main contributions:
\begin{itemize} 
\item A new approach for indoor robot navigation that offers two main advantages:  
i) A navigation scheme centered in action selection that is less prompt to 
localization errors, and ii) A semantically rich and compact representation that facilitates path 
planning and interaction 
with human users. 
\item A working system that uses a virtual environment to show the viability of 
the proposed approach. 
    \item An open source implementation that will be available online to foster 
further research in this area. \end{itemize}

\input{relatedWork}

\section{\label{Approach}PROPOSED METHOD}

\iffalse
\begin{figure*}
    \centering
{\small
\begin{tabular}{ccccc}
    \multirow{3}{*}{\includegraphics[height=3.1cm]{map_landmarks_and_office.pdf}}
    &\multirow{3}{*}{\includegraphics[height=3.1cm]{map_paths.pdf}}
    &\multirow{3}{4.2cm}{\includegraphics[width=0.25\textwidth]{map_graph.pdf}}
    & \textbf{Map Graph} & \textbf{Plan}: \\
    & & & \nr{place}{behavior}{place}\ & S:O1 G:O8\\
    & & & \nr{O1}{oor}{C1r} & \nr{O1}{oor}{C1r}\\
    & & & \nr{O1}{ool}{C1l} & \nr{C1r}{cf}{H1}\\
    & & & \nr{O1}{ooc}{EO3} & \nr{H1}{chr}{C2r}\\
    & & & \nr{EO3}{io}{O3} & \nr{C2r}{cf}{EO6}\\
    & & & \nr{C1r}{cf}{H1} & \nr{EO6}{io}{O6}\\
    & & & \ldots & \\
    & & & \nr{O16}{oor}{C4r} & \\
    & & & \nr{O16}{ooc}{EO14} & \\
    (a) & (b) & (c) & (d) & (e)
\end{tabular}
}
%\includegraphics[width=\textwidth,height=7cm]{twoApproaches.pdf} 
\caption{Need a good caption and a better figure.}\label{twoApproaches}
\end{figure*}
\fi

\begin{figure}
    \centering
{\small
\begin{tabular}{cc}
\includegraphics[height=3.5cm]{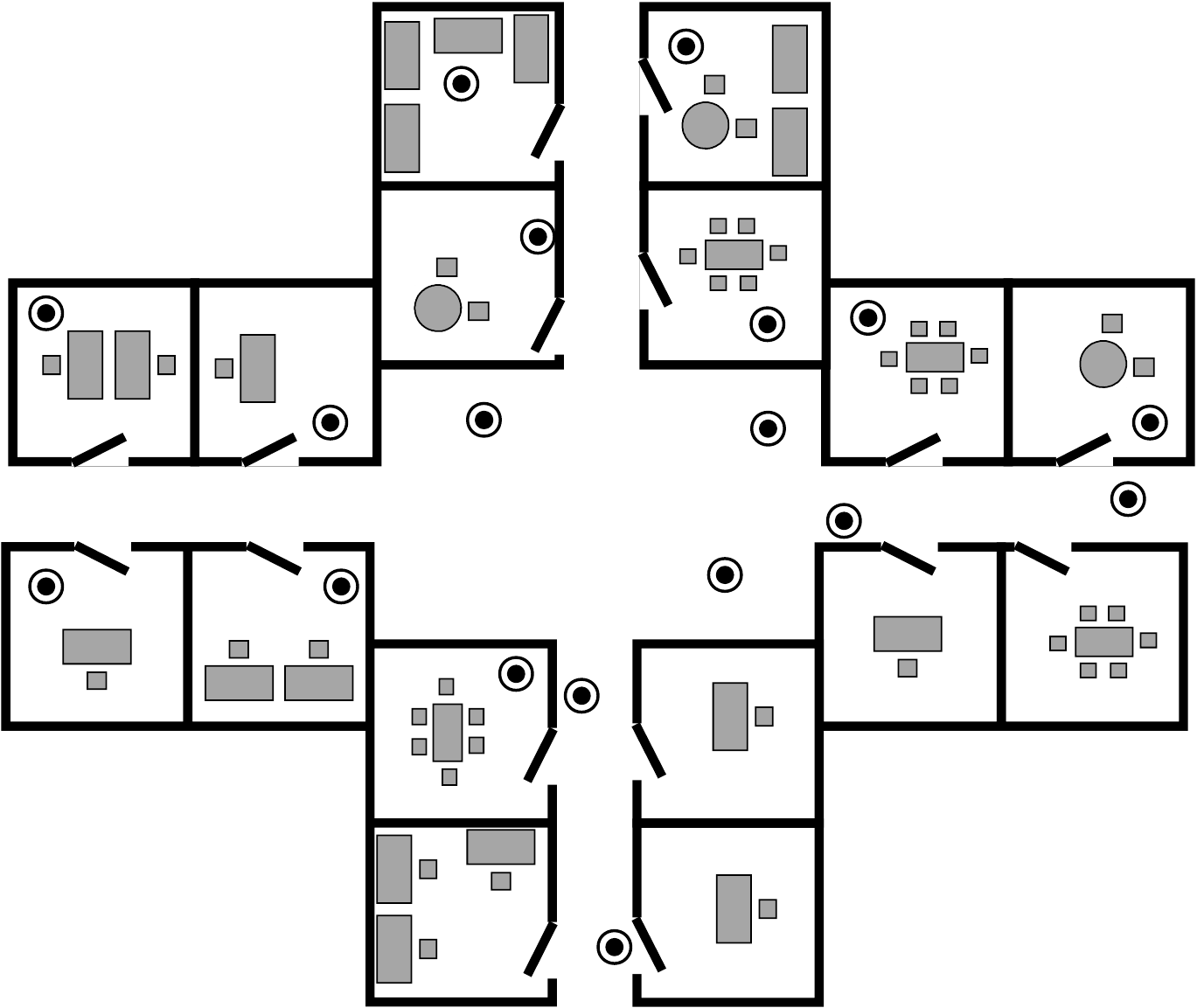}
    & \includegraphics[height=3.5cm]{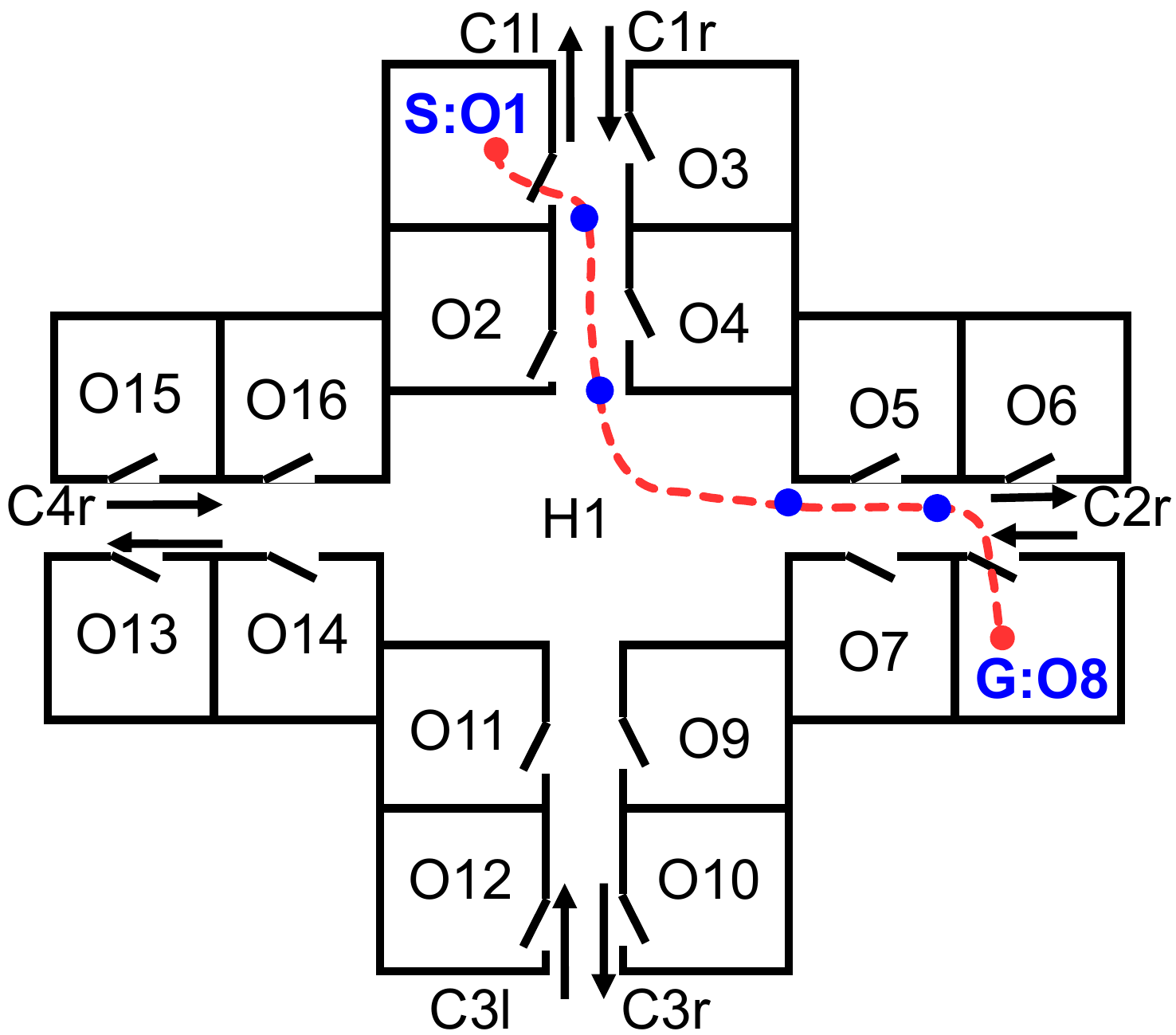}\\
(a) & (b) \\
\multicolumn{2}{c}{\includegraphics[width=0.8\columnwidth]{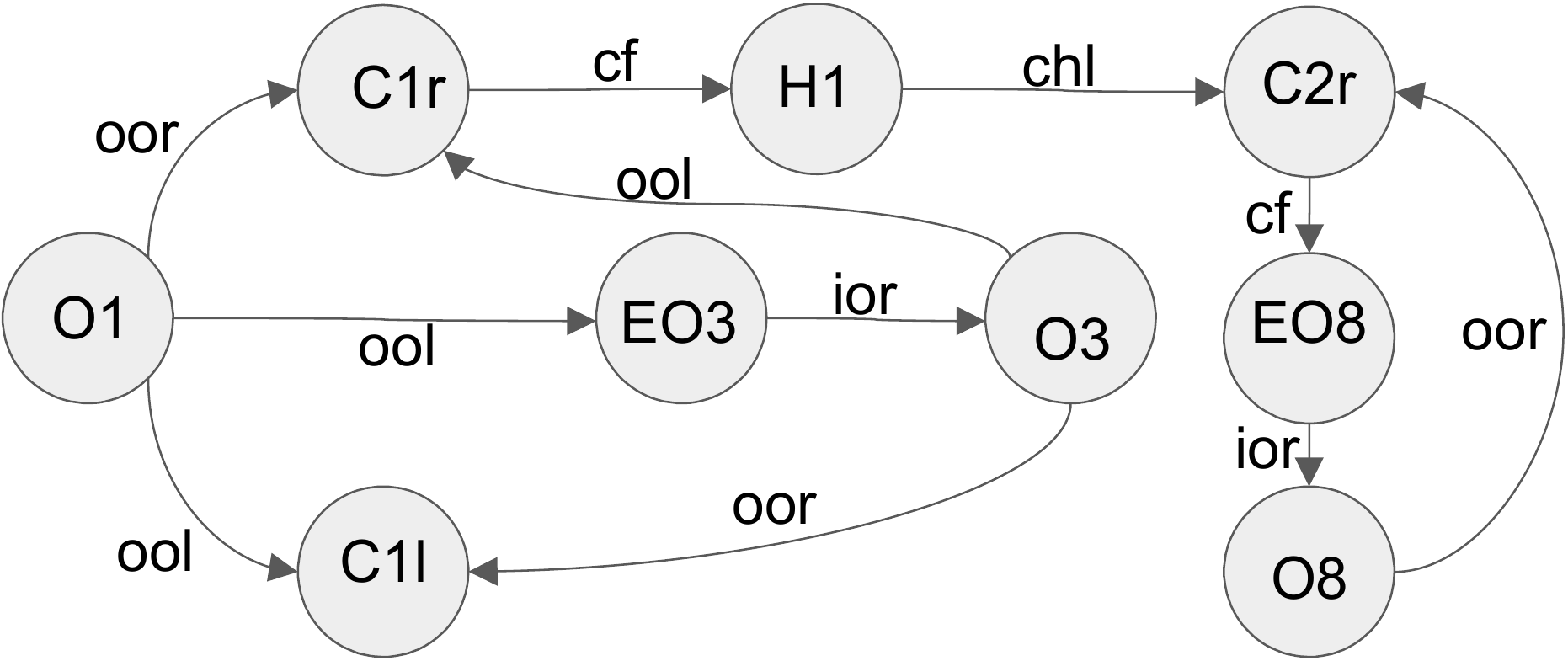}}\\
\multicolumn{2}{c}{(c)}\\

\textbf{Map Graph} & \textbf{Plan}: \\
\nr{place}{behavior}{place}\ & S:O1 G:O8\\
\nr{O1}{oor}{C1r} & \nr{O1}{oor}{C1r}\\
\nr{O1}{ool}{C1l} & \nr{C1r}{cf}{H1}\\
\nr{O1}{ooc}{EO3} & \nr{H1}{chl}{C2r}\\
\nr{EO3}{ior}{O3} & \nr{C2r}{cf}{EO8}\\
\nr{C1r}{cf}{H1} & \nr{EO8}{ior}{O8}\\
\ldots & \\
\nr{O16}{oor}{C4r} & \\
\nr{O16}{ooc}{EO14} & \\
(d) & (e)\\
\end{tabular}
}
%\caption{(a) Combined metric and topological map representations incorporate information
%about geometry and landmarks of the environment.
%(b) Instead, we use a graph-based representation derived from key locations in the environment such 
%as offices (O1, O2, \ldots), corridors ($C_{1l}$, \ldots), etc.
%(c) A partial map graph for this example environment. (d) Alternatively, we can represent this map 
%as a set of triplets of the form \nr{place}{behavior}{place}.
%(e) When we need to navigate from O1 to O8 (as depicted in red in (b)), we can derive a behavioral 
%plan by following appropriate nodes in our map graph.}
%\label{twoApproaches}
%\end{figure}
\caption{(a) A typical geometric representation used to implement robot navigation in an office 
building enviroment. This consists of a metric map (grey structures) augmented with information 
about relevant visual landmarks (circles with a dot in the center). (b) Instead, we use a 
graph-based representation derived from 
the identification of key semantic locations in the environment, such as offices (O1, O2, \ldots), 
halls (H1), and corridors ($C_{1l}$, \ldots). (c) A partial graph for the environment depicted in 
b) (see Appendix for notation details). (d) This graph can be 
represented as a set of triplets of the form 
\nr{place}{behavior}{place}.
(e) To navigate from O1 to O8 (as depicted in red in (b)), one can derive a 
behavioral plan by following appropriate links (triplets) in the graph.}
\label{twoApproaches}
\end{figure}

Our key innovation is the use of a representation of the environment given by a set of perceptual 
and navigational behaviors. Furthermore, similarly to \cite{Bojarski:NVIDIA:2016}, these behaviors 
are robustly learned using deep learning techniques, specifically supervised imitation learning. 
This scheme marks a departure from current state-of-the-art SLAM techniques that mainly 
rely on metric and landmark-based representations of the environment. 

Figure \ref{twoApproaches} shows a hypothetical case that helps to illustrate our representation as 
well as to highlight its main differences with respect to current approaches for indoor robot 
navigation. Figure \ref{twoApproaches}-a shows an example of a typical map representation used by 
current robots to navigate in an office environment. This consists of a metric map (grey structures) 
augmented with information about relevant visual landmarks (circles with a black dot in the center), 
while the metric representation encodes the configuration of relevant structures and objects, the 
landmark-based topological representation increases localization robustness by encoding the 
appearance and relative position of specific visual landmarks. As a relevant fact, the key 
component 
of this hybrid representation is the encoding of geometrical information about the environment. 
Figure \ref{twoApproaches}-b represents the same environment, but using the labels of meaningful 
semantic locations, such as offices (O1, O2, \ldots), halls (H1), and corridors 
($C_{1l}$, \ldots). Using these labels, Figure \ref{twoApproaches}-c shows an 
excerpt of an instantiation of the 
proposed 
graph representation for the office environment considered in a) and b). As shown in Figure  
\ref{twoApproaches}-d, this graph can 
also be represented as a set of navigational rules or triplets that take the 
form: \nr{place}{behavior}{place}\footnote{See Appendix for 
notation details.}. As an example, office $O_1$ is 
connected by a 
navigational behavior \textit{out-of-office-right} (\textbf{oor}) to  the right 
direction of corridor $C_1$ ($C_{1r}$), leading to the triplet: 
\nr{O_1}{oor}{C_{1r}}. Similarly, corridor $C_1$ direction right ($C_{1r}$) 
is connected by a navigational behavior \textit{corridor-follow} (\textbf{cf}) 
to hall $H_1$, leading to the triplet: \nr{C_{1r}}{cf}{H_1}. Notice that 
given the difference between traversing a corridor through the left or 
right direction, we introduce a different place label to distinguish
these cases. 

As a relevant observation, in contrast to metric and landmark-based approaches that focus on 
encoding geometry, the key component of the proposed representation is the encoding of semantic 
information about the environment. Where we understand semantic as a set of high-level skills, or 
behaviors, that allow a robot to successfully navigate in a man-made environment. As a relevant 
advantage, the proposed representation facilitates the use of planning techniques. As an example, 
Figure \ref{twoApproaches}-e illustrates a valid plan to go from office $O_1$ to office $O_8$, which 
can be expressed as: Leave office $O_1$ and take corridor $C_1$ direction right, 
\nr{O_1}{oor}{C_{1r}}. Follow corridor $C_{1r}$ until reaching hall $H_1$, \nr{C_{1r}}{c_f}{H_1}. 
Then, cross hall $H_1$ and take to the left to joint corridor $C_{2r}$, \nr{H_1}{chl}{C_{2r}}. 
Next, follow corridor $C_{2r}$ until the entrance of office $O_8$, \nr{C_{2r}}{cf}{EO8}. Finally, 
enter office $O_8$ on the right, \nr{EO8}{ior}{O_8}. The example also highlights 
a relevant additional advantage of the proposed approach, that is, its affinity with the way humans 
navigate in indoor environments, a property that can largely facilitate the interaction of robots 
with human users. 

The selection and robust implementation of suitable navigational and perceptual behaviors is 
the key element behind the proposed approach. Indeed, the 
viability of the proposed approach resides on being able to robustly 
implement behaviors such as mastering how to leave an 
office or how to traverse a corridor. As we describe next, DL
techniques allow us to acquire such capabilities using training data.

\subsection{\label{Behaviors} Navigational Behaviors.}
Following \cite{Pomerleau:1993} and \cite{Bojarski:NVIDIA:2016}, we use 
supervised imitation learning (SIL) as our primary strategy to implement the 
intended navigational behaviors. As a main advantage, SIL takes advantage of a 
direct observation of how an expert solve similar problems, avoiding a brute 
force or blind exploration of the state space 
\cite{Levine:RSS:2017}. Similarly to \cite{Bojarski:NVIDIA:2016}, we implement a SIL scheme based 
on deep Convolutional Neural Networks (CNNs) and visual information.

\begin{figure}
\begin{center}
\includegraphics[width=4.0cm]{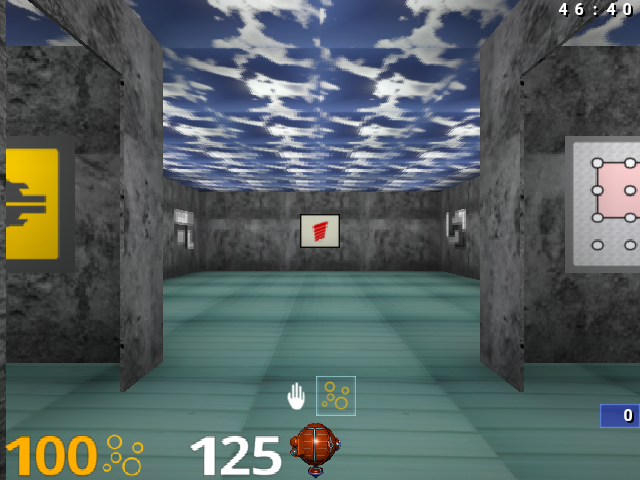} 
\hspace{0.1cm}
\includegraphics[width=4.0cm]{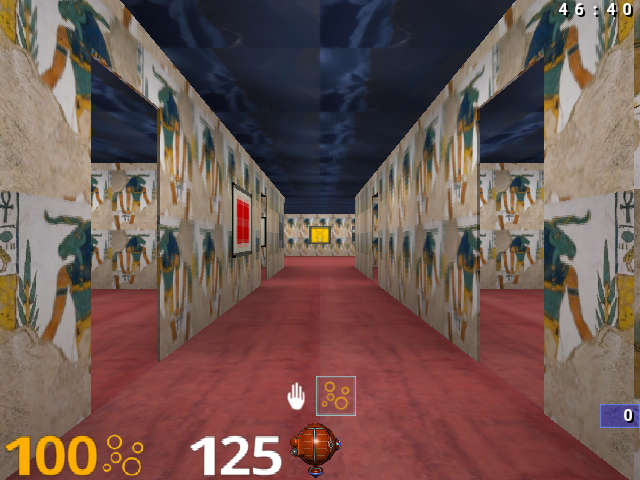} 
\end{center}
\caption{Examples of the virtual environments generated by the DML framework 
\cite{DeepMindLab:2016}.}
\label{Fig:DeepMind}
\end{figure}

As a relevant requirement, the implementation of DL models needs large amounts 
of training data, in our case, direct observation of expert behaviors. Several strategies can be 
used to 
collect this data. As an example, \cite{Bojarski:NVIDIA:2016} and 
\cite{droneTrails:2016} directly gather training data by recording a first 
person camera view along with control actions while a human executes a specific behavior. As an 
alternative, it is possible 
to gather training data by registering the behavior of a virtual agent while it 
navigates in 3D reconstructions of natural environments, such as 
\cite{IroArmeni:EtAl:CVPR-2017}, or in photorealistic renderings of virtual 
environments, such as \cite{Yuke:EtAl:2017}. As a proof of concept, in  this 
work, we use the virtual environments provided by the research framework 
DeepMind Lab (DML) \cite{DeepMindLab:2016}. DML provides 3D virtual environments 
that are built on top of a game engine. In particular, this framework provides a 
set of tools to collect first-person-view images 
and control actions while an agent traverses a simulated environment. Figure 
\ref{Fig:DeepMind} shows images that illustrate the 3D environments provided by 
DML.
 
\begin{figure}
\begin{center}
\includegraphics[width=8.5cm]{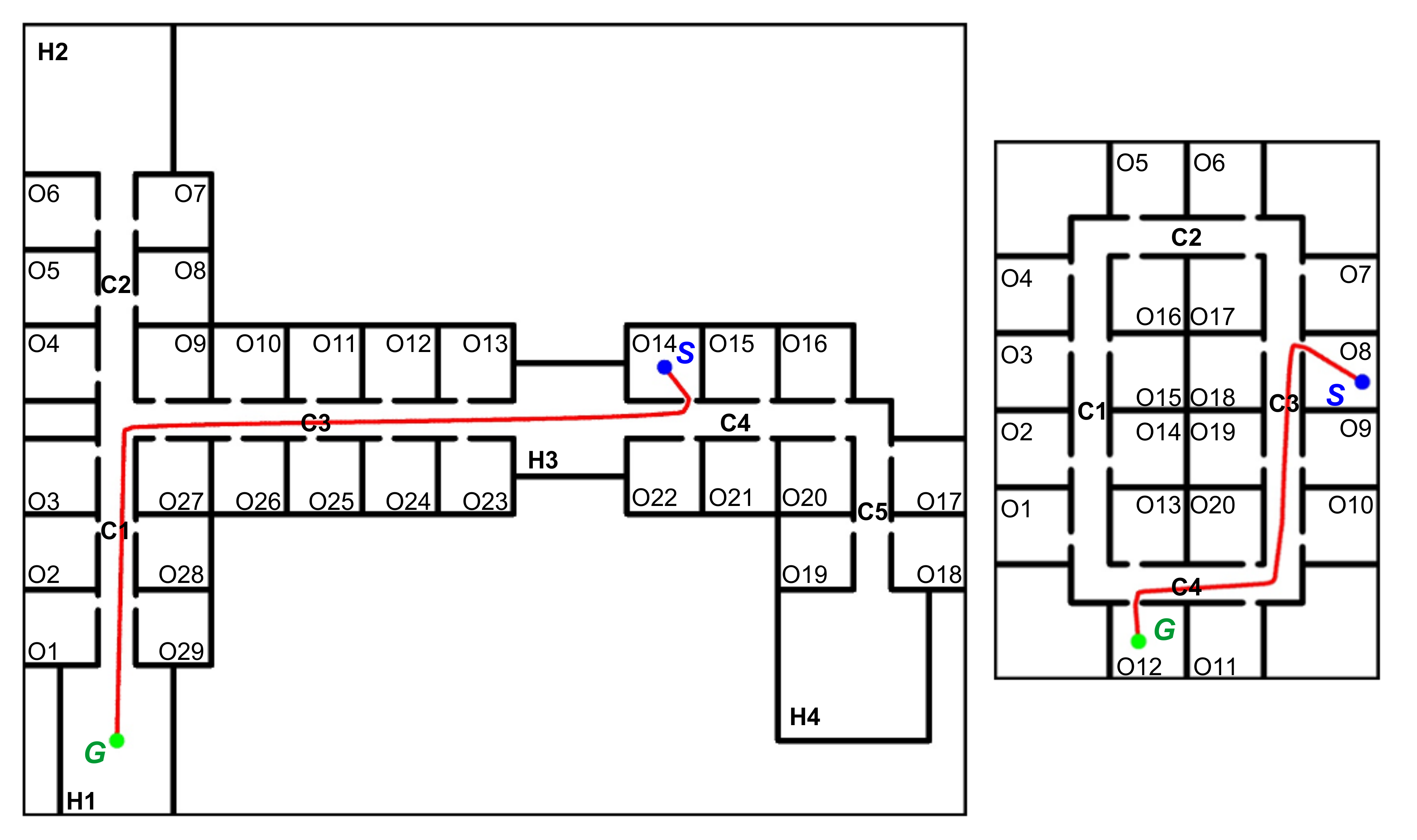}
\end{center}
\caption{Examples of office building floor maps randomly generated to gather 
training data to test the proposed approach. Given a random start location S, a 
traditional path planner (ex. RRT* \cite{RRT:2011}) is used to simulate the path that an 
expert will follow to reach a goal location G.} 
\label{Fig:Maps}
\end{figure}

To generate data corresponding to the navigational behavior of an expert, we 
augment the DML framework with two main tools. First, we implement a tool that 
generates random 2D floor plans that simulate office buildings. 
Specifically, these floor plans are composed of a random arrangement of 
structural components, mainly office rooms, halls, and corridors. Figure 
\ref{Fig:Maps} shows examples of configurations generated by this tool. 
Additionally, using a 2D floor map as input, we implement a second tool that interacts with DML to 
generate a 3D rendering of 
the floor map, leading to simulated environments as the ones shown in Figure \ref{Fig:DeepMind}.

Using the previous tools we are able to generate the expert behaviors. 
Specifically, using a 2D floor plan, we exploit 
the optimality of traditional path planning techniques 
\cite{Latombe:Planning:1991}, to simulate the route that an expert would 
follow when 
executing each of the intended behaviors (ex. \textit{entering-an-office}, 
\textit{corridor-follow}, etc). After calculating an optimal path, a 
virtual agent executes the route in the virtual world, recording at the same 
time training data corresponding to visual images and control actions faced by 
the expert. Additionally, following \cite{droneTrails:2016} we account for 
perturbations with respect to the expert routes by adding jitter to the 
trajectory of the virtual agent (views in a +5$^{\circ}$ and -5$^{\circ}$ 
orientations). The previous scheme provides large 
sources of training data that can be used to learn the intended behaviors. 

To obtain the plan to navigate from a start (S) to a goal (G) location, we use 
the 
implementation of the planner RRT* \cite{RRT:2011} provided by OMPL 
\cite{OMPL:2012}. Figure \ref{Fig:Maps} shows examples of random starts (S) and 
goal locations (G) as well as the path generated by RRT*. It is important to 
mention that to generate a path that favors a route that follows the center of 
the corridor, we use a C-space configuration to increase the thickness of the 
walls in 
the map \cite{Latombe:Planning:1991}. Furthermore, to closely resemble the path used by 
a human, we use a cubic spline to smooth the 
resulting path provided by the planner. 

\subsection{\label{Sec:DL-Models} DL Models.}
In our implementation, we consider two types of behaviors i) 
\textit{Reactive} and ii) \textit{Memory-based}. 

\noindent \textit{i) Reactive behaviors:} correspond to behaviors that 
implement a direct mapping from sensing to action, i.e., they do not use an explicit 
internal representation of the world. In our case, this is a direct mapping from visual input 
to robot motor control. We envision these behaviors as highly general. As an 
example, a corridor following behavior should be able to successfully operate in 
a wide variety of indoor environments. As a relevant feature, our 
navigational behaviors are not purely reactive, but they also incorporate an 
intended goal. As an example, a navigational behavior to cross a hall also 
takes into consideration the direction that the robot will take 
after leaving the hall. 

\begin{figure}
\begin{center}
\includegraphics[width=6cm]{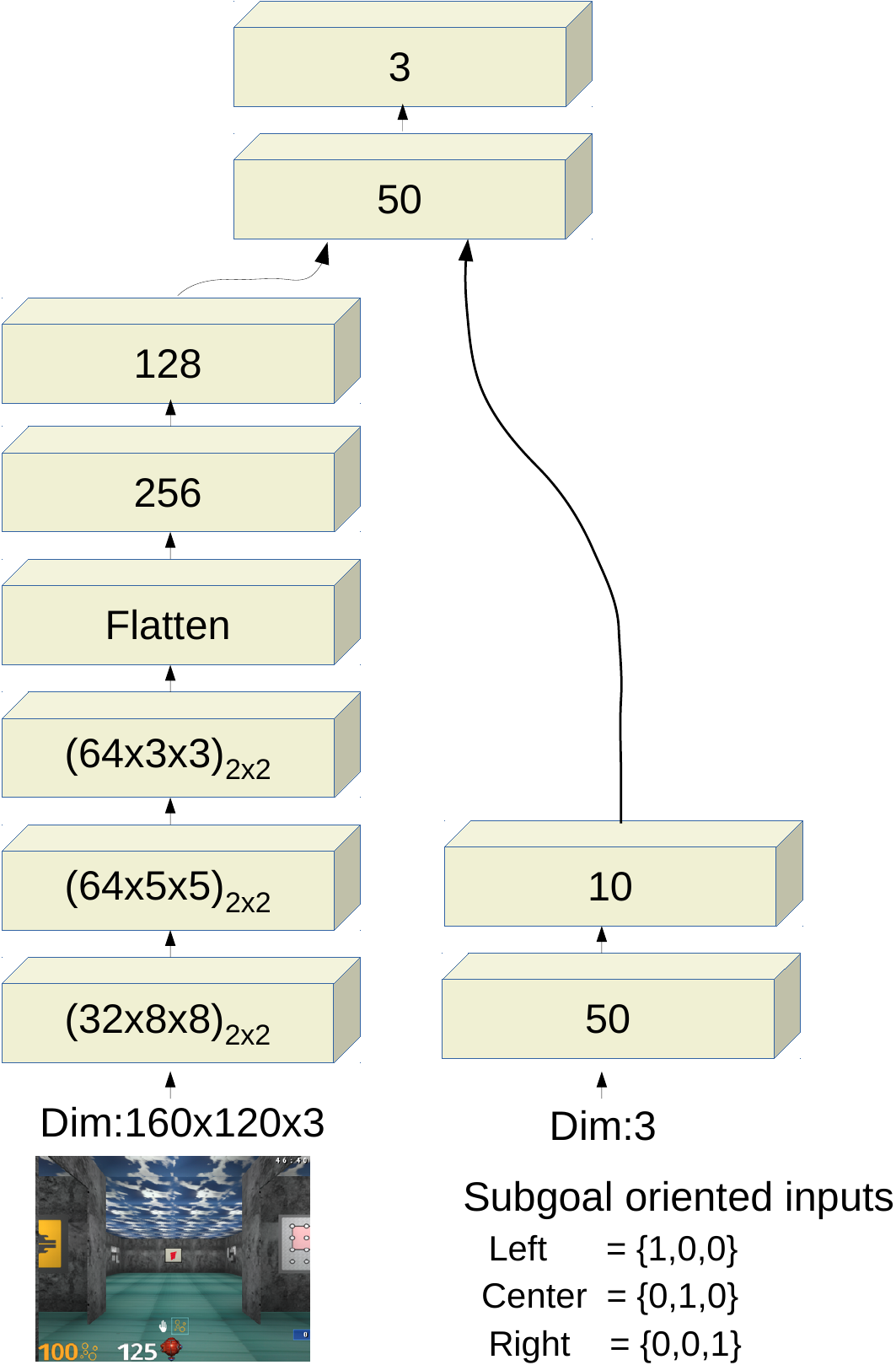} 
\end{center}
\caption{We implement \emph{reactive behaviors} by mapping directly from visual observations to 
actions using Convolutional Neural Networks (CNNs). Multiple behaviors may be implemented with a 
single CNN by using an additional subgoal oriented input. As an example the \emph{out-of-office} 
behavior can be turned into \emph{out-of-office-right} or \emph{out-of-office-left} depending 
on the subgoal oriented input.} \label{Fig:CNN}
\end{figure}

Figure \ref{Fig:CNN} shows the CNN used to learn the reactive behaviors. 
The 
input is given by the current view of the robot and a one-hot 3D vector that provides 
additional information about subgoals related to the current execution 
of the behavior, such as leaving a hall using the exit of the left. Given that our robot moves in a 
flat 
floor, the output of the CNN is given by a 3D vector indicating the desired 
translational ($x,y$) and rotational ($\theta$) velocities. These outputs are 
given in robot 
world coordinates (Y axis perpendicular to the robot motion direction).

\noindent \textit{ii) Memory-based behaviors:} correspond to behaviors 
that use an explicit internal representation, or memory, to recognize previously 
visited places. In contrast to our reactive behaviors which learn navigational 
skills that are valid for different environments, our memory-based behaviors encapsulates specific 
knowledge that is only valid for a particular environment. As an example, a memory-based behavior 
will have the ability to recognize a 
specific 
office or hall. In particular, memory-based behaviors play a relevant role to 
signal transitions between reactive behaviors. As an example, while following a corridor, the 
detection of a particular office entrance will signal the switching between a 
\textit{corridor-follow} and \textit{input-office} behaviors.  

\begin{figure}
\begin{center}
\includegraphics[width=6.5cm]{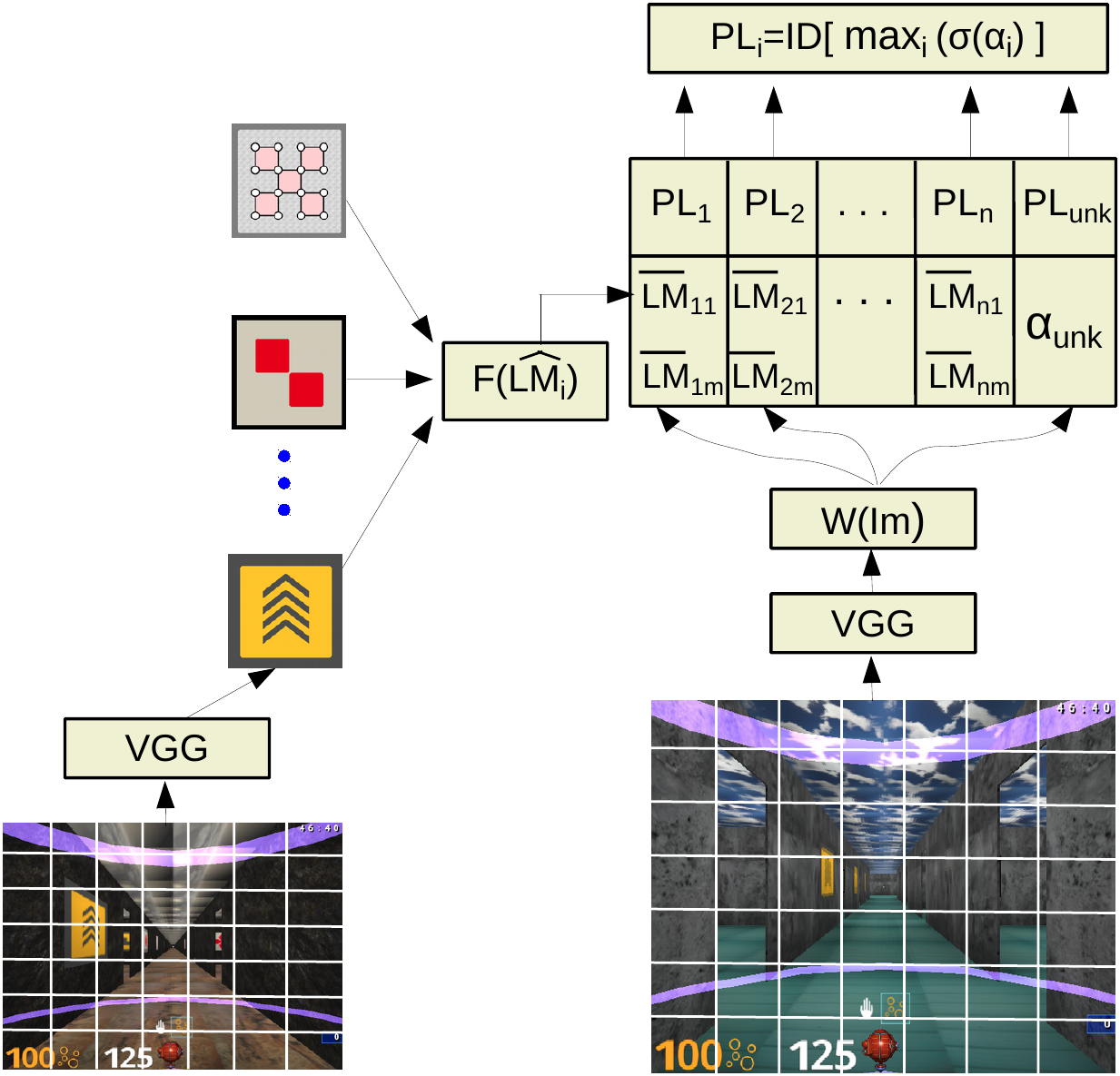} 
\end{center}
\caption{An example of a memory-based behavior. In particular, this behavior uses a 
pretrained CNN and a memory module to identify 
environment places associated to the detection of specific visual landmarks, 
see main text for detail.} \label{Fig:MemNW}
\end{figure}

Figure \ref{Fig:MemNW} illustrates our method to 
implement a memory-based behavior that allows to identify specific places of the 
environment. For simplicity, when we generate each 
environment we associate a specific visual 
landmark to each meaningful place. Therefore, place recognition is reduced to landmark 
detection, 
where these landmarks are given by pictures hanging from the walls (see figures 
\ref{Fig:DeepMind} and \ref{Fig:CrossHall}).  

Specifically, let's $LM=\{LM_1, \dots, LM_n\}$ be 
the set of $n$ known visual landmarks. During training, for each landmark 
$LM_i$ we obtain a set of $m$ visual descriptors $\widehat{LM}_{ij}$, $j \in [1,\dots,m]$, by 
applying a 
pre-trained CNN to a set of $m$ images that are known to contain landmark $LM_i$. 
Each visual descriptor is given by the last 
pooling layer of 
the pre-trained VGG-Net model \cite{VGG-2015}, which provides a feature map 
corresponding to 7x7 image regions. During training, we only 
consider the image region that most overlap with the known position of the 
landmark. Afterwards, we apply to the resulting $m$ descriptors $\widehat{LM}_{ij}$ a trainable 
bilinear function $F(\cdot)$ that provides embbedings $\overline{LM}_{ij}$. We store these 
embeddins as keys of $nxm$ memory locations. The 
addressable value of each of these 
memory locations corresponds to a place-ID $PL_i$, indicating the place associated 
to the respective landmark (each landmark appears only once in the 
environment). Additionally, we also consider a default place-ID $PL_{unk}$ to handle cases where 
the input image does not contain a valid visual landmark. Consequently, our method for place 
recognition resembles the association mechanism used by key-value memory networks 
\cite{MemNWs:2015}.

At detection time, we apply to the current view of the robot $Im$,  the same pretrained VGG-Net 
model described above, obtaining a set of 7x7 descriptors. 
Afterwards, we apply to each descriptor a trainable bilinear function
$W(\cdot)$ obtaining encodings $Im_l$, $l \in [1,\dots,7x7]$, that we use to address the memory 
locations with information about places. In particular, we use cosine distance to score the 
similarity between each image region
descriptor and the encoding of the visual landmarks stored in memory. Using the 
resulting scores, we estimate a 
probability $p(PL_i)$ of being at each possible place $PL_i$. Specifically, we calculate 
$p(PL_i)$ as follows:

\begin{align} \label{Eq:MemNw}
p(PL_i) = \sigma(\alpha_i) 
\end{align}
where:
{\small
\begin{align*}
\alpha_i= \sum_{l=1}^{7x7}\sum_{j=1}^m <Im_l, LM_{i,j}> \;\; \mbox{and} \;\;
\sigma(\alpha_i)= \frac{e^{\alpha_i}}{\sum_{i} e^{\alpha_i} + e^{\alpha_{unk}}}
\end{align*}
}

\noindent $<\cdot,\cdot >$ denotes cosine distance, $\sigma(\cdot)$ is a softmax 
function, and $\alpha_{unk}$ corresponds to a constant 
coefficient associated to the default place-ID $PL_{unk}$. 
This coefficient 
can be considered as a threshold over the probability of accepting a valid 
landmark detection. After training, we replace the softmax function in Eq. \ref{Eq:MemNw} for a 
hard maximum operation, reporting as detected place $PL_{i}$ associated to the 
coefficient $\alpha_i$ with highest value.

\section{\label{Sec:Experiment}Experiments}
As we mention, we test the proposed approach using simulated environments that 
resemble the configuration of office building spaces. As illustrated in Figure 
\ref{Fig:Maps}, our current implementation considers 3 types of 
office building spaces: offices, corridors, and halls. In terms of 
behaviors, while it is possible to explore automatic 
techniques to identify them \cite{Justin:EtAl:2017}\cite{NeuralTuring:2015}, 
for simplicity, we manually preselect a set of behaviors that can meet the 
navigational requirements of the intended environment. Specifically, we 
implement the following behaviors:

\begin{itemize}
 \item Corridor-Follow (\textbf{cf}): a reactive navigational behavior that leads  
the way of the robot as it navigates through a corridor. The input to this 
behavior is the current view of the robot (RGB image).

 \item Out-Office (\textbf{oo}): a reactive navigational behavior that leads the 
way of the robot as it leaves an office. The input to this behavior is the 
current view of the robot and a 3D vector encoding the direction that 
the robot should take after leaving the office: \textit{right} or
\textit{left}.  

\item Enter-Office (\textbf{io}): a reactive navigational behavior that leads  the 
way of the robot as it navigates through a corridor and enters to the next 
office on its way. The input to this behavior is the 
current view of the robot and a 3D vector encoding the direction 
that the robot should take to enter the next office:  \textit{right} or 
\textit{left}.

\item Cross-Hall (\textbf{ch}): a reactive navigational behavior that leads  the 
way of the robot as it crosses a hall. The input to this behavior is the 
current view of the robot and a 3D vector encoding the direction 
that the robot should follow to leave the corridor: \textit{right}, 
\textit{left}, or \textit{center}.

\item Change-Corridor (\textbf{cc}): a reactive navigational behavior that leads the 
way of the robot as it faces an intersection to follow a new 
corridor. The input to this behavior is the 
current view of the robot and a 3D vector encoding the direction of the next 
corridor: \textit{right}, \textit{left}, 
or \textit{center}.

\item Place Detection (\textbf{pd}): a perceptual behavior that identifies the 
type of place where the robot is currently located. Possible categories 
are: office, corridor, hall.

\item Landmark-based Place Detection (\textbf{lmpd}): a memory based perceptual behavior that 
identifies places using the detection of specific LMs in the environment. Possible categories are a 
set 
of $n=80$ available visual LMs, as described in Section \ref{Sec:DL-Models}. 
\end{itemize}

\subsection{\label{Sec:Result1}Behavior training}

Input images consist of 160x120 pixels. We test using RGB and gray level 
images (BW), both schemes converge to suitable models, but BW achieves faster 
convergence. Therefore, we use BW images to train all behaviors, except \textbf{lmpd} that 
uses RGB images. We normalize each input image by subtracting the mean and 
dividing by the standard deviation, both estimated over the training set. 
We use Adam as the optimization tool, keeping the recommended 
parameter values \cite{Adam:2014}. We use 
randomly sampled batches of 256 instances. Initial learning rate is set to $10^{-4}$. 
Also, we apply batch-normalization \cite{BatchNormalization:2015} after every 
convolutional layer (except the last one).     

\begin{figure}
\begin{center}
\includegraphics[width=8.7cm]{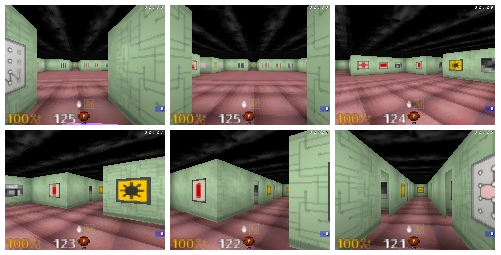}
\end{center}
\caption{Sequential set of frames corresponding to samples of a 
training instance of behavior 
cross-hall. From left to right, the frames illustrate the robot entering, then 
crossing, and finally leaving the hall using the 
right side direction. } 
\label{Fig:CrossHall}
\end{figure}

Table \ref{Table:Training} provides details about the datasets 
used to learn each behavior. To collect this data, we use virtual 
environments to render expert trajectories consisting of the robot executing 
each behavior. As an example, Figure \ref{Fig:CrossHall} shows a set of 
frames corresponding to a training instance used to fit the behavior 
\textit{cross-hall}. In this case, the robot crosses the hall, following afterwards the right side corridor. As shown in Table 
\ref{Table:Training}, the size of each dataset is variable and mostly depends on 
the expected training complexity of each behavior. 

\begin{table}[h]
\begin{center}
\resizebox{\columnwidth}{!}{
\begin{tabular}{|l|c|c|c|c|c}
\toprule
\multicolumn{5}{c}{\textbf{Navigational Behaviors}} \\
\bottomrule
  & Train: \#paths & Train: \#images & Test: \#trials & 
Accuracy\\
\bottomrule
\textbf{cf} & 150 & 46131 & 100 & 100\% \\
\textbf{io} & 8000 & 221734 & 100 & 100\% \\
\textbf{oo} & 8000 & 259937 & 100 & 67\% \\
\textbf{oo-b} & 8000 & 259937 & 100 & 96\% \\
\textbf{ch} & 3000 & 248950 & 100 & 100\% \\
\textbf{cc} & 4000 & 123653 & 100 & 100\% \\

\toprule
\multicolumn{5}{c}{\textbf{Perceptual Behaviors}} \\
\bottomrule
  & - & Train: \#images & Test: \#images & 
Class. Acc.\\
\bottomrule
\textbf{pd} & - &12945 & 1500 (Imgs) & 98.2\% \\
\textbf{lmpd} & - & 38459 & 1500 (Imgs) & 96.7\% \\
\toprule
\end{tabular}
}
\end{center}
\caption{Details of the datasets used to train each behavior as well as the 
accuracy achieved on an independent test set. For navigational behaviors, we 
measure accuracy as the number of trials where the virtual robot successfully 
complete the goal of the behavior, ex. robot successfully leaves an office using 
the intended departing direction. For perceptual behaviors, we measure accuracy 
as the successful classification rate.}
\label{Table:Training}
\end{table}

In terms of navigational behaviors, Table \ref{Table:Training} considers accuracy 
as the percentage of trials where the robot successfully completed 
the intended behavior, ex. crossing a corridor successfully using the intended 
departing direction. In terms of perceptual behaviors, accuracy 
is measured as the percentage of images where the model outputs a correct 
classification. For most behaviors, results in Table 
\ref{Table:Training} indicate a robust operation. The exception is 
behavior \textit{out-of-office} (\textbf{oo}) that only succeed in 67\% of the trials. This poor 
performance is mainly due to the random scheme used to initialize the robot 
position, which causes that the robot occasionally starts in a position directly 
facing a wall. In these cases, one will expect that the robot consistently rotates 
in one direction, searching for the position of the door to leave the 
office. However, the memoryless property of the CNN that is used to implement 
the behavior can not handle this capability. As a result, robot usually 
starts wandering close to the wall entering a deadlock situation. While it 
is possible to use a recurrent model to implement this behavior, we decide to 
restrict the initial position of the robot in such a way that it always starts facing the 
initial direction of the path of the expert, which always points towards the 
office door. Introducing this simplification, we obtain behavior \textbf{oo-b} 
that is able to achieve high robustness.

\subsection{\label{Sec:Results2} Behavior Integration}
Next, we test the performance of the proposed approach to solve navigational 
tasks that require the integration of several behaviors. To conduct this 
experiment, we randomly generate 100 maps. For each map, we randomly select 
10 navigational tasks consisting on moving between 2 different offices. This 
situation is similar to the case depicted in Figure \ref{twoApproaches}-b, 
where the robot needs to navigate from office $O_1$ to office $O_8$. For 
simplicity, we directly extract the graph representation from the known 
configuration of the environment. In a more general case, this map can be 
inferred by a guided presentation of the environment to the robot or by an 
initial unsupervised exploration phase. 

Table \ref{Table:Test} shows our results. Accuracy is measured as
the percentage of missions where the robot successfully reaches the intended 
goal location. As expected, the robot is 
able to successfully finish its mission in a significant proportion of the 
cases. Furthermore, the average length of 8.3 activations of navigational behaviors to complete 
each mission shows that the proposed approach is effective to control the selection and 
switching of behaviors. In terms of failure cases, a manual analysis reveals 
that these are mainly due to two problems. First, during robot navigation, the 
landmark detection behavior (\textbf{lmpd}) occasionally fails to detect the landmark 
that identifies the entrance to the goal office. As a result, the robot misses 
the activation of behavior \textit{enter-office} (\textbf{io}), traversing its last corridor 
until its end, to finally wonder without destination. This problem 
is related to the use of a first person camera view that is always pointing to 
the current heading direction of the robot. As a result, occasionally, this 
camera is not pointing in a suitable direction missing the detection of a 
relevant landmark. A second failure case is due to the presence of repeated 
textural patterns in the virtual environment. In particular, the visual 
appearance of the view of a door just before leaving an office is highly similar 
to the view of the front door after leaving the office, as a result in some 
occasions the robot takes a wrong action, such as turning to face a 
hypothetical corridor before leaving the office. In practice, we 
believe that in a real implementation the previous two problems can be solved, 
and they do not represent a major limitation to the viability of our 
approach. 

\begin{table}[h]
\begin{center}
\begin{tabular}{|l|c|c|c|c|c}
\hline 
\#Maps & \#Missions/map & \#Missions & Av. Steps/mission & 
Acc.	\\
\hline\hline
100   & 10 & 1000 & 8.3 &  81.2\% \\ 
\hline
\end{tabular}
\end{center}
\caption{Performance of the proposed approach when robot needs to complete 
a mission consisting of navigating between 2 different offices. Accuracy is 
measured as the percentage of missions where the robot successfully reaches the 
intended goal location. We also report the average number of steps to complete 
a mission, which is equivalent to the average number of navigational behaviors needed to 
reach the goal location.}
\label{Table:Test}
\end{table}

\section{\label{Conclusions}CONCLUSIONS}
We present a new approach for indoor robot navigation that uses a graph 
representation to integrate perceptual and navigational 
behaviors. This approach offers an alternative view that departs from 
current methods and representations used to solve the robot navigation problem. 
As we discuss, current approaches focus on encoding geometry, as a way to 
facilitate robot localization and from there to select suitable actions to achieve
navigational goals. In contrast, the proposed approach focuses on encoding 
suitable behaviors that directly guide the robot to complete its navigational 
goals. As a result, the proposed method offers two main advantages. On one hand, 
it leads to a robot navigational scheme that is less prompt to localization errors 
(ex. due to changes in the environment). On other hand, it relies on a 
semantically rich and compact representation of the world that facilitates the 
interaction with human users as well as the application of planning 
techniques.

An important aspect of the proposed approach is the determination of a set of 
behaviors that allow the robot to complete its navigational goals. 
In this work we show that, in simple scenarios, manual selection can 
lead to a working solution. Further research is needed to devise 
automatic methods that can deal with more complex situations. Recent work on 
modular visual reasoning \cite{Justin:EtAl:2017} appears as an attractive option to 
fulfill this goal.

A relevant observation of the experimental validation is the limitation of pure reactive behaviors 
to successfully deal with the wide variety of situations faced by a purposeful robot. The inclusion 
of an extra orientation input helps us to discriminate between cases where a specific subgoal 
governs the overall execution of a specific behavior, such as leaving a hall using the left or right 
direction. However, the failure of the behavior \textit{out-of-office} indicates the need to include 
models that can handle short-term memories.

The proposed approach opens several interesting research avenues. As an 
example, the current implementation separates the mapping and planning steps, 
we believe that it is possible to face these problems under a common learning 
framework. Recent work on the so-called Neural Turing Machine 
\cite{NeuralTuring:2015} appears as an attractive option to joint perception, 
representation, and control under a common framework. Also, the 
integration of the proposed representation with a natural language model is 
an interesting research avenue \cite{Zang:EtAl:2018}. This can take advantage of the high-level 
semantic behind 
the proposed representation, leading to a more natural way to implement robot-human interaction. 

Finally, it is important to notice that while our current implementation introduces 
certain simplifications to facilitate the operation of navigational and perceptual behaviors, we 
believe that the fundamental ideas behind the 
proposed approach still hold in a more general case. This is in part supported 
by current success exhibited by deep learning techniques, which have demonstrated 
exceeding robustness to deal with complex perception and control 
tasks in natural environments. In this sense, a pending challenge is 
the implementation of the proposed approach using a real robot.

\section{\label{Sec:Discussion}APPENDIX: NOTATION}

\begin{table}[h]
\begin{center}
{\footnotesize
\begin{tabular}{|l|l|}
\hline
Code & Description \\
\hline 
$\mathbf{ool}$& out of office, take left \\ 
$\mathbf{oor}$& out of office, take right \\ 
$\mathbf{cf}$& follow-corridor \\ 
$\mathbf{iol}$& enter office to the left \\ 
$\mathbf{ior}$& enter office to the right \\ 
$\mathbf{chs}$& cross hall, continue straight \\ 
$\mathbf{chl}$& cross hall, take left\\ 
$\mathbf{chr}$& cross hall, take right\\ 
$\mathbf{ccc}$& change corridor, straight\\ 
$\mathbf{ccl}$& change corridor to the left\\ 
$\mathbf{ccr}$& change corridor to the right\\ 
\hline
\end{tabular}
}
\end{center}
\caption{Summary of the navigational behaviors that we use in our implementation.}
\label{Table:Notation}
\end{table}

%%%%%%%%%%%%%%%%%%%%%%%%%%%%%%%%%%%%%%%%%%%%%%%%%%%%%%%%%%%%%%%%%%%%%%%%%%%%%%%%

%%%%%%%%%%%%%%%%%%%%%%%%%%%%%%%%%%%%%%%%%%%%%%%%%%%%%%%%%%%%%%%%%%%%%%%%%%%%%%%

%%%%%%%%%%%%%%%%%%%%%%%%%%%%%%%%%%%%%%%%%%%%%%%%%%%%%%%%%%%%%%%%%%%%%%%%%%%%%%%%

\bibliographystyle{IEEEtran}
\bibliography{AsaReferences}

\end{document}

%% file: relatedWork.tex
\section{\label{relatedWork}RELATED WORK}
There is an extensive literature discussing the problem of autonomous robot 
navigation, we refer the reader to \cite{Kostavelis:Gasteratos:2015} for 
a recent review. Here, we focus our discussion on indoor environments, where the robot does not 
have access to a global positioning system 
(GPS). Similarly, in terms of the abundant literature about DL models 
\cite{DeepLearning:Science:2015}, we focus our discussion on applications 
related to visual indoor navigation.

Inspired by \cite{Walter:1950}, behavioral approaches to robot navigation gain 
high popularity during the 80s and 90s. In particular, the subsumption 
architecture becomes one of the best known approaches \cite{Brooks:1986}. Under 
this scheme, behaviors are implemented as finite state machines that are 
combined through mutual suppression mechanisms (subsumption). As a key 
underlying principle, subsumption states that the world is its own best 
representation, therefore, there is not need to keep an explicit internal 
representation of the environment. Instead, a robot can directly uses its current estimation 
of the state of 
the world to trigger the opportunistic activation of suitable behaviors. Subsumption based robot 
navigation shows remarkable 
success to emulate reactive behaviors observed in biological systems, such as 
insects \cite{Brooks:Genghis:1989}. It also achieves some degree of success to 
perform simple tasks in natural scenarios, mainly office 
buildings \cite{Horswill:1993}. 

In contrast to subsumption approaches, later research has stressed the need to 
keep an internal representation of the world in the form of a map of the environment. In 
particular, the mapping and 
localization problems have been usually handled in conjunction, leading to the Simultaneous 
Localization and Mapping or SLAM problem  
\cite{Thrun:2006}. Initial SLAM approaches are based on the Kalman Filter (KF) and 
its extensions \cite{Smith:Self:Cheeseman:1990}. Limitations of the Gaussian 
assumption in KF techniques lead to non-parametric approaches based on the 
Particle Filter \cite{Montemerlo:EtAl:2002}. Later, limitations of 
non-parametric approaches to scale to large environments lead to Graph-Based 
methods \cite{Grisetti2010a}. Under this scheme, the challenge is 
to find a configuration of the nodes in the graph that is maximally consistent with the 
observations of the environment, which implies solving a complex optimization 
problem. Several works have proposed highly efficient solutions 
\cite{Thrun:GraphSlam:2006}\cite{Dellaert-GraphSlam:2006} that are able to 
operate in real time for large environments ($>10^8$ landmarks) \cite{ 
Kaess:Dellaer:Isam:07icra}\cite{OlsonSlam:09}. While SLAM approaches make 
intensive use of range and odometry sensors, advances in computer vision 
techniques facilitate the development of the so-called Visual-SLAM (V-SLAM) 
approach based on visual sensors \cite{vSLAM-Review:12}. As an 
example, \cite{CumminsIJRR08} presents FAB-MAP, a SLAM approach based on the 
detection of 2D visual keypoints. Extensions using RGB-D sensors 
have also been proposed \cite{huang2011isrr}.

The previous SLAM techniques rely on low level structural information which 
limits their generality and robustness. Moreover, they lack of a suitable 
semantic representation that facilitates the interaction with human users. This has 
motivated recent extensions that incorporate high level semantic 
information \cite{Kostavelis:Gasteratos:2015}, such as the ability to recognize objects 
\cite{GalvezEtAl:2015}\cite{Kostas:ICRA:2017}, scenes 
\cite{Espinace:EtAl:2013}, or navigational structures \cite{Reid:2014}. As an 
example, \cite{Kostas:ICRA:2017} proposes a Graph-Based SLAM approach that 
includes the detection of semantic objects, such as chairs and doors. As 
expected, results in \cite{Kostas:ICRA:2017} indicate that the understanding of the environment at 
a higher level of abstraction helps to increase 
navigational robustness.

Closer to our ideas, there are recent works on 
robot navigation using a DL approach, mainly deep reinforcement learning (DRL) 
techniques. 
\cite{Gupta:EtAl:2017} presents the so-called Cognitive Mapper and Planner 
(CMP), a system that integrates mapping and planning using a joint learning 
scheme. Given visual and self-motion information, this system learns a policy to 
achieve generic goals, such as go to a target location or find an object of 
interest. Similarly, \cite{Yuke:EtAl:2017} presents a system that learns a 
policy to move a robot to a given view point position of the environment. In 
other words, 
the goal is given by an image of the target location. Interestingly, both works, 
\cite{Gupta:EtAl:2017} and \cite{Yuke:EtAl:2017}, use virtual environments to 
obtain a first navigational model that is then refined using real data. In 
contrast to our work, \cite{Gupta:EtAl:2017} and \cite{Yuke:EtAl:2017} do not 
learn an explicit representation of 
the environment and they are not based on a behavioral approach.

In terms of works that use DL to learn an explicit map of the environment, \cite{Ruslan:Nav:17} 
presents the so-called Neural Map, a system that keeps a 2D representation of the environment 
similar to a 2D occupancy grid \cite{Elfes:1989}. As a relevant feature, instead of storing binary 
information in each map cell (busy/empty), they store a feature vector provided by a DL model. In a 
related approach, \cite{Honglak:Nav:2016} presents a navigation approach based on a temporal 
extension of Memory Networks \cite{MemNWs:2015}. As a result, they are able to improve 
navigational performance with respect to pure reactive approaches that do not consider an explicit 
external memory. \cite{DeepMind:Nav:2017} proposes a DRL approach that improves robot navigation by 
considering auxiliary tasks, such as depth prediction and loop closure classification. As a result, 
they are able to improve navigational performance with respect to approaches that only 
consider the navigational task. In contrast to our approach, none of the works above use an 
internal representation based on navigational behaviors.

In the biological side, the use of an internal representation of the world has been supported by 
highly influential studies that demonstrate the use of an explicit representation of the 
world by the navigational systems of rodents and humans 
\cite{OKeefe:1978}\cite{Moser:2005}. Here, we also advocate for the use of an explicit 
internal representation of the world. As a main novelty, we propose 
an approach that deeply connects this representation to the execution of goal oriented behaviors.